\documentclass[lettersize,journal]{IEEEtran}
\usepackage{amsmath,amsfonts}
\usepackage{algorithmic}
\usepackage{algorithm}
\usepackage{array}
\usepackage[caption=false,font=normalsize,labelfont=sf,textfont=sf]{subfig}
\usepackage{textcomp}
\usepackage{stfloats}
\usepackage{url}
\usepackage{verbatim}
\usepackage{graphicx}
\usepackage{cite}
\usepackage[table,xcdraw]{xcolor}

\usepackage{colortbl}
\usepackage[table]{xcolor}
\usepackage{multirow}
\usepackage{listings}%
\usepackage{xcolor}%
\usepackage{manyfoot}%
\usepackage{booktabs}%
\usepackage{color}
\usepackage{amssymb}

\begin{document}

\title{An Angular-Temporal Interaction Network for Light Field Object Tracking in Low-Light Scenes}

\author{
    Mianzhao Wang,
    Fan Shi,
    Xu Cheng,
    Feifei Zhang, 
    and Shengyong Chen,~\IEEEmembership{Senior Member,~IEEE}
\thanks{This work was supported by National Natural Science Foundation of China (NSFC) (62272342, 62020106004, and T2422015), Natural Science Foundation of Tianjin Municipality (CN) (23JCJQJC00070), Tianjin Municipal Science and Technology Program (CN) (24PTLYHZ00320). Corresponding author: Fan Shi.}
\thanks{
Mianzhao Wang, Fan Shi, Xu Cheng, Feifei Zhang, Shengyong Chen are with the Engineering Research Center of Learning-Based Intelligent System (Ministry of Education), the key Laboratory of Computer Vision and System (Ministry of Education), the School of Computer Science and Engineering, Tianjin University of Technology, Tianjin, 300384, China (e-mail: wmz@stud.tjut.edu.cn, shifan@email.tjut.edu.cn, xu.cheng@ieee.org, feifeizhang@email.tjut.edu.cn, sy@ieee.org). 
}

}

\markboth{Journal of \LaTeX\ Class Files,~Vol.~14, No.~8, August~2021}%
{Shell \MakeLowercase{\textit{et al.}}: A Sample Article Using IEEEtran.cls for IEEE Journals}


\maketitle

\begin{abstract}
High-quality 4D light field representation with efficient angular feature modeling is crucial for scene perception, as it can provide discriminative spatial-angular cues to identify moving targets. However, recent developments still struggle to deliver reliable angular modeling in the temporal domain, particularly in complex low-light scenes. In this paper, we propose a novel light field epipolar-plane structure image (ESI) representation that explicitly defines the geometric structure within the light field. By capitalizing on the abrupt changes in the angles of light rays within the epipolar plane, this representation can enhance visual expression in low-light scenes and reduce redundancy in high-dimensional light fields. We further propose an angular-temporal interaction network (ATINet) for light field object tracking that learns angular-aware representations from the geometric structural cues and angular-temporal interaction cues of light fields. Furthermore, ATINet can also be optimized in a self-supervised manner to enhance the geometric feature interaction across the temporal domain. Finally, we introduce a large-scale light field low-light dataset (R8LUT) for object tracking. Extensive experimentation demonstrates that ATINet achieves state-of-the-art performance in single object tracking. Furthermore, we extend the proposed method to multiple object tracking, which also shows the effectiveness of high-quality light field angular-temporal modeling.
\end{abstract}

\begin{IEEEkeywords}
Light field, Object tracking, Low-Light, Epipolar plane images.
\end{IEEEkeywords}

\section{Introduction}

\IEEEPARstart{L}{ight} describes the multi-angular variation of light, encompassing both its propagation direction and intensity at each point in free space \cite{TPAMIdisparityestimation},\cite{TPAMIreconstruction},\cite{zhang2016light}. 
Existing tracking models \cite{ARTrack},\cite{OSTrack} are primarily developed based on RGB camera techniques, which use a main lens to capture light intensity converging onto a single point from multiple directions, thereby forming a 2D image \cite{ATOM},\cite{MAE}. As these methods rely solely on intensity-based representations, their effectiveness is significantly compromised in complex low-light environments, where intensity information is heavily attenuated. Consequently, 2D RGB images offer limited visual cues, leading to degraded tracking performance.
Conversely, light field cameras acquire structured 4D light field data in a single shot using an array of micro-lenses, simultaneously recording spatial-domain intensity and angular-domain directional information \cite{chen2022deep}. This rich representation significantly enhances the appearance and geometric structural description of objects, even under low-light conditions.
Recently, light field has shown great promise in various vision tasks, including depth estimation \cite{wang2022disentangling}, \cite{strecke2017accurate}, saliency detection \cite{TPAMIreconstruction}, \cite{wang2022lfbcnet}, and semantic segmentation \cite{cong2023combining}. 
Motivated by these advantages, in this paper we aim to develop a light field object tracking framework that operates across both angular and temporal dimensions to improve target discrimination in low-light scenes.

In a light field imaging system, the captured structured light field video can be defined as $L(u, v, x, y)$, where $(u, v)$ represents the angular coordinate system, and $(x, y)$ represents the spatial coordinate system \cite{li2023opal},\cite{zhou2023beyond}. Fixing two angular coordinates implies observing the scene from a specific angular domain, resulting in a general sequence of two-dimensional images \cite{cong2023combining}. On the other hand, fixing one angular coordinate and one spatial coordinate means gathering pixels from different angular domains, forming a sequence of epipolar plane images (EPIs) \cite{han2021novel}. EPIs provide a structured representation of the high-dimensional light field, revealing the geometric structural properties of the light field \cite{strecke2017accurate}, \cite{srinivasan2017shape}. Therefore, the light field encapsulates not only appearance and texture details in the spatial domain but also conveys geometric structural information across the angular and temporal domains, contributing to a more comprehensive understanding in low-light scenes. However, due to the high-dimensional structure of the light field, the angular domain tightly couples redundant appearance cues while sparsely distributing geometric structural cues. Although EPIs showcase the distribution of geometric structural cues in the angular domain by rearranging multi-view pixels \cite{MAC}, these representations still include redundant appearance cues, limiting the expressive capacity of light fields in the angular domain. Additionally, this redundant coupling necessitates the design of task-specific angular modules to explore geometric structural cues in the angular domain \cite{wang2022visual}, which in turn reduces the learning performance of the model. Therefore, decoupling the geometric structural cues of the light field in the angular domain remains a key challenge.

After obtaining geometric structural cues from the light field, the next crucial step is modeling temporal cues for light field object tracking. Over the past decades, various relational modeling methods have been proposed \cite{ATOM}, \cite{CenterTrack}. These methods can be categorized into two classes: the first category directly connects adjacent frames and inputs them into neural networks, implicitly generating features that amalgamate motion and appearance information \cite{OSTrack}, \cite{MixFormer}. The second category introduces appearance models, establishing contextual connections for images, and further utilizes temporal modules based on CNN or Transformer architectures to explicitly capture the temporal information in video sequences \cite{Trades}, \cite{ToMP}. While these methods have excelled in 2D object tracking tasks, they primarily focus on exploring spatio-temporal correlations based on dense pixel data. For sparse light field structural information, these methods might inadvertently aggregate some information from non-geometric structures of light field into the feature representation, thereby weakening the light field’s ability to distinguish between targets and backgrounds. Thus, modeling temporal features of light fields requires not only consideration of inter-frame and intra-frame correlations in the angular domain but also effective aggregation of sparse geometric structural clues across angular-temporal dimensions. Additionally, current motion perception frameworks typically undergo supervised pre-training on large-scale video datasets \cite{ToMP},\cite{ByteTrack}, followed by fine-tuning for downstream video tasks. However, due to the limited availability of light field datasets, relying solely on supervised learning is insufficient for effectively modeling the temporal features of light fields. Therefore, how to establish angular-temporal correlation for light field video and ensure effective learning is another key challenge.

In this paper, we propose a novel light field epipolar-plane structure image (ESI) representation. This representation explicitly delineates the geometric structure points within the light field by capitalizing on the abrupt changes in the angles of light rays within the epipolar plane, thereby enhancing the exploration of motion cues in the low-light scenes. We then propose an angular-temporal interaction network (ATINet) for light field object tracking, designed to excavate rich discriminative features across angular-temporal dimensions. The proposed ATINet not only models inter-frame and intra-frame correlations in the angular domain but also efficiently gathers sparse geometric structure cues across angular-temporal dimensions through adaptive selection. Additionally, we introduce a self-supervised loss designed to optimize angular-aware representations for ATINet. This promotes interactions among geometric structural features in the temporal domain, thereby facilitating effective feature learning to support temporal matching. Next, we establish a large-scale dataset (R8LUT) for light field single object tracking and multiple object tracking tasks, providing 173 and 26 light field videos respectively, each complete with bounding boxes and ESI representations. Finally, we evaluate the proposed ATINet on the light field single object tracking (SOT) task and extend the proposed method to the multiple object tracking (MOT) task with simple adjustments. Experimental results on these tasks demonstrate the competitive performance of our method, validating its effectiveness in low-light scenes. Our main contributions are:
\begin{itemize}
	\item We propose a novel light field ESI representation that explicitly describes the geometric structure points within the light field by exploiting the abrupt properties of light rays within the epipolar plane, thereby enhance the expression of visual cues in low-light scenes. Compared to existing light field representations, the light field ESI removes high-dimensional redundant information and eliminates the need for designing complex angular modules.
	\item  We introduce an angular-temporal interaction network (ATINet) for light field object tracking, which learns angular-aware representations from both geometric structural and angular-temporal interaction cues of light fields to address complex low-light scenes. Additionally, ATINet can be optimized using a self-supervised approach to improve the interaction of geometric features over the temporal domain.
	\item We construct a large-scale light field dataset (R8LUT) for object tracking, comprising multiple complex objects carefully arranged in low-light scenes. Additionally, we validate the proposed method on light field SOT tasks and further extend it to MOT tasks, which have not been explored by previous methods. Extensive experiments demonstrate that our method achieves state-of-the-art performance on these tracking benchmarks.
\end{itemize}

\section{Related work}

\subsection{Light Field for Computer Vision}
Light field imaging simultaneously captures spatial and angular information from multiple directions in a single shot, providing explicit scene cues widely utilized in various computer vision tasks \cite{han2021novel}. 
Early studies primarily focused on depth estimation from light fields to exploit their unique spatio-angular information\cite{zhang2016light}.
Recently, other visual tasks based on light fields have also garnered significant attention \cite{wang2022lfbcnet}. Zhang et al. \cite{zhang2021geometry} explored spatial complementary information and disparity correlation among multiple views in salient object detection. 
Cong et al. \cite{cong2023combining} introduced a light field semantic segmentation network that efficiently fuses complementary information within the light field. 
Meanwhile, Wang et al. \cite{wang2022visual} proposed a light field single object tracking framework, which integrates the spatial-angular information to enhance tracking performance. 
Although considerable advancements have been made in computer vision for light fields, current research has not yet explored motion relation modeling, thereby limiting their generalizability to video scenarios. In this paper, we explicitly delineate geometric structure points in light field and introduce an angular-temporal interaction network for light field video.

\subsection{Object Tracking}
Object tracking involves initializing the target's state in the first frame and tracking its subsequent movements. 
In the history of object tracking, Discriminative Correlation Filters (DCFs) emerged as the predominant paradigm \cite{ECO}. Subsequent extensions of DCF-based trackers \cite{PrDiMP},\cite{DiMP} have consistently demonstrated excellent performance across multiple tracking benchmarks. 
With the advancement of deep learning, Siamese neural networks have emerged as a powerful tool in tracking. Bertinetto et al. \cite{SiamFc} introduced SiamFC, designed to train the network from end to end. Building upon the foundation of SiamFC, researchers have conducted extensive studies, including backbone architectures \cite{li2019siamrpn++}, online model update \cite{zhang2019learning}, and target state estimation \cite{wang2021unsupervised}. 
Furthermore, transformers have been employed in object tracking and have garnered great attention \cite{OSTrack,MixFormer,SwinTrack}. These methods extract features through a single-stream or dual-stream encoder and utilize multi-head attention mechanisms to establish temporal correlations across multiple frames. 
Although object tracking have achieved significant success, current methods that rely solely on appearance models experience performance degradation in low-light scenes. Our proposed model addresses this issue by effectively learning light field temporal cues, demonstrating its benefits in object tracking tasks.

\subsection{Self-supervision Video Representation Learning}
To circumvent the costly and time-intensive process of video data annotation, several self-supervised methods have been developed to extract video features from large-scale unlabeled videos \cite{Survey_self_Supervised_1,Survey_self_Supervised_5}. 
Self-supervised video representation learning trains neural networks using the objective functions of pretext tasks, leveraging the inherent structure of the data as a supervisory signal. Vondrick et al. \cite{vondrick2018tracking} introduced a self-supervised learning method through video colorization to capture temporal information, leveraging the natural temporal consistency of colors to colorize grayscale videos by copying colors from reference frames. 
Recently, inspired by the great success of self-supervised learning in NLP \cite{Devlin2019BERTPO}, masked image modeling (MIM) \cite{MAE} have demonstrated promising results in video representation learning. 
Sun et al. \cite{sun2023masked} captured long-term and fine-grained motion cues from sparse video inputs by reconstructing motion trajectories. Meanwhile, Gupta et al. \cite{gupta2023siamese} focused on object motion by maintaining the invariance of past frames while masking the majority of patches in future frames to learn object-centered representations. While such methods have demonstrated superior performance in 2D video, their application to learning light field video representation remains unexplored. In this paper, we constructed a light field self-supervised loss to extract angular-temporal features.

\begin{figure*}[t!]
	\centering
	\includegraphics[width=0.85\linewidth]{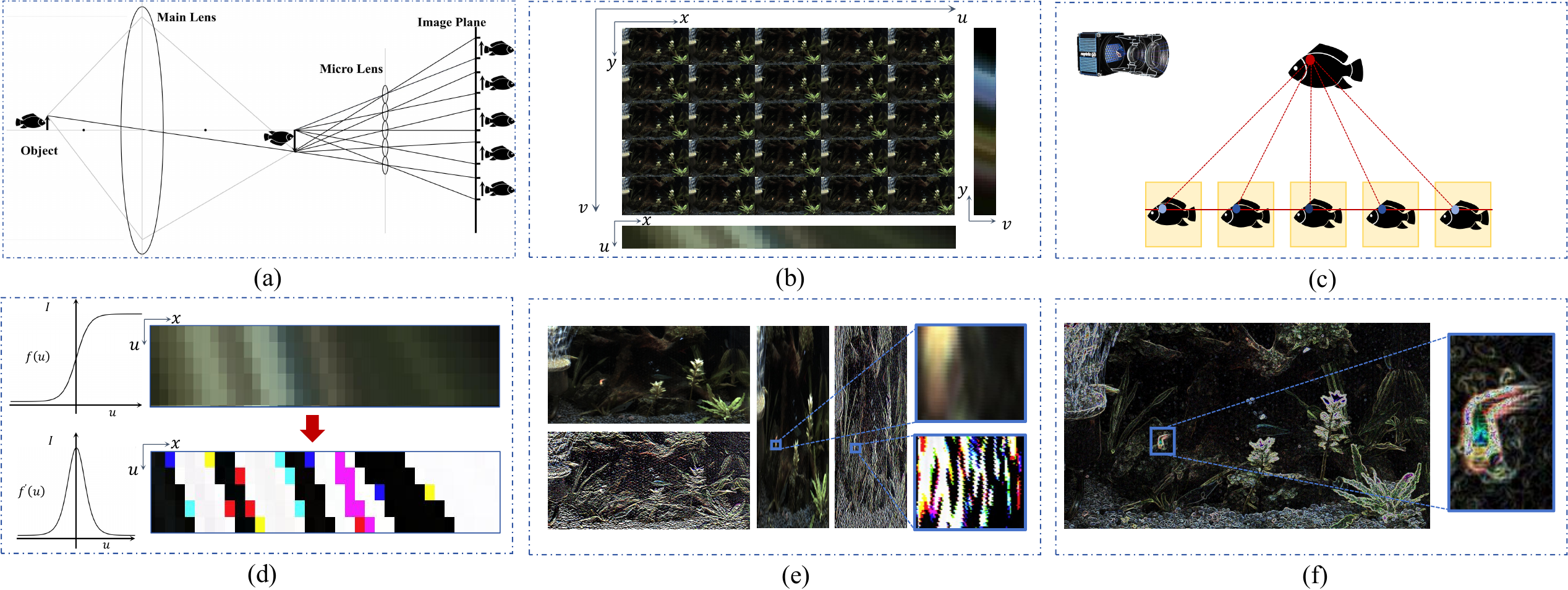}
	\caption{Illustrations of light field representations:
(a) Light field imaging system.
(b) Light field multi-view and EPI representation.
(c) Changes in viewpoints within the light field.
(d) Calculation of geometric structure points.
(e) Projection of all geometric structure points.
(f) Generated light field ESI representation.}
	\label{Figure1}
\end{figure*}

\section{Methods}
\subsection{Light Field Representation}
The 4D light field, denoted as $L(u, v, x, y) \in \mathbb{R}^{U\times V\times W\times H}$ and represented by the two-plane model, can be transformed into alternative representations \cite{li2023opal},\cite{zhou2023beyond}, such as EPI images $\hat{L}_H(u, x) \in \mathbb{R}^{U\times W} $ or $\hat{L}_T(v, y) \in \mathbb{R}^{V\times H}$, as shown in Fig. \ref{Figure1}(a-b). This representation with line structure attributes can be seamlessly integrated with traditional CNNs to extract intricate spatial-angular cues for scene understanding. While beneficial in some respects, the excessive redundancy of appearance information within EPI significantly hampers the light field’s ability to express geometric structural properties in the angular domain. Effectively capturing the geometric structure of a light field in the angular domain is a key question.

Within a light field, the direction of light rays undergoes notable variations near the boundaries of objects due to their distinct refraction across different depth of field, as shown in Fig. \ref{Figure1}(c). This phenomenon effectively delineates the geometric attributes of the light field. Consequently, this paper focuses on the abrupt changes of light rays, aiming to explicitly depict the geometric structure properties of the light field and eliminate unnecessary redundancy, thereby facilitating the exploration of temporal motion cues. Given that the directional properties of light rays are represented within the line structure of EPI, identifying locations with the maximum gradient in the angular dimension corresponds to pinpointing the spots where variations in light rays are most pronounced, as shown in Fig. \ref{Figure1}(d). We name these specific points as geometric structure points within the light field. Based on the analysis above, we introduce a novel representation of light field, known as epipolar plane structure image (ESI). Next, we elucidate the construction process of ESI.

Given a candidate horizontal EPI image $\hat{L}_H(u,x)$ of $L(u,v,x,y)$, the gradient of a point along angular dimension can be described in conjunction with other viewpoints. That is, each EPI can reconstruct the geometric structure points through gradient computations. Specifically, for a horizontal EPI image, the gradient is formulated as follows:
\begin{equation}
\label{EPI1}
\hat{L}_{u}^{'}=[\alpha \hat{L}(u) + \beta \hat{L}(u+d_{l}) + \gamma \hat{L}(u+d_{r})  ]+\eta_{u}
\end{equation}
where $d_{l}$ and $d_{r}$ represent the left and right step sizes in the angular dimension, respectively. $\alpha$, $\beta$, $\gamma$ are coefficients, with $\eta_{u}$ representing the consistency error. According to Taylor series expansion analysis, $\hat{L}_{u}^{'}$ can be approximated as follows:
\begin{equation}
\label{EPI2}
\hat{L}_{u}^{'}=\frac{d^{2}_{r} \hat{L}(u+d_{l})+(d^{2}_{l}-d^{2}_{r})\hat{L}(u+d_{l})-d^{2}_{l}\hat{L}(u-d_{r})}{d_{l}d_{r}(d_{l}+d_{r})}
\end{equation}
Since the step size between angles in EPI is equal to the distance between adjacent micro-lens array centers (i.e., $d_{l}$=$d_{r}$),  can be simplified as:
\begin{equation}
\label{EPI3}
\hat{L}_{u}^{'}=\frac{\hat{L}(u+d)-\hat{L}(u-d)}{2d}
\end{equation}
Here, we define $d=1$, $u=U/2$, which detects abrupt changes in light rays based on the central view and its adjacent angular space. Therefore, we can search for geometric structure points by examining the first-order derivatives on all EPIs. Next, as shown in Fig. \ref{Figure1}(e), we project these generated geometric structure points onto the plane to form horizontal EPI gradient image $S_H(x, u, y)\in \mathbb{R}^{W \times U\times H} $. Following the same method, the vertical EPI gradient image $S_T(v, x, y) \in \mathbb{R}^{V \times W \times H} $ can also be obtained. Finally, we calculate the amplitude of the horizontal and vertical EPI gradient images to form the final light field ESI, the process is formulated as follows:
\begin{equation}
\label{EPI4}
S=\sqrt[2]{S_H(x,u=U/2,y)^{2}+S_T(v=V/2,x,y)^{2}}
\end{equation}
The constructed light field ESI $S\in \mathbb{R}^{H\times W}$ is shown in Fig. \ref{Figure1}(f). This representation explicitly delineates the geometric structure points within the light field by detecting the maximum values of the first-order derivatives in the angular space of the EPI. Therefore, we can observe that in low-light scenes, the light field ESI clearly represents the contours of the target.

\begin{figure*}[t!]
	\centering
	\includegraphics[width=0.95\linewidth]{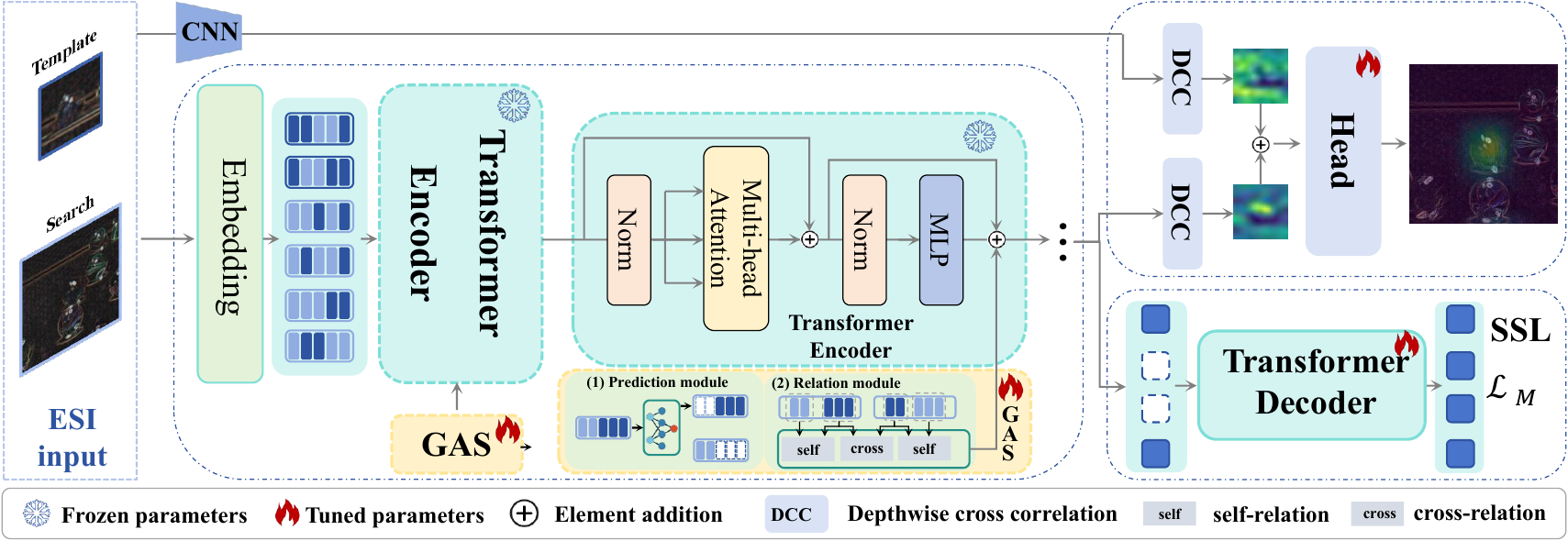}
	\caption{An overview of our ATINet framework, which is a hybrid architecture tracker that employs Depthwise cross correlation to localize targets. It uses the ESI representation as input and integrates the proposed baseline with GAS as the feature extractor with a CNN backbone. During the training phase, only the GAS, the head networks, and decoder are fine-tuned. Additionally, the proposed SSL is used exclusively during the training phase.}
	\label{framework}
\end{figure*}

\subsection{Light Field Angular-Temporal Interaction Network}
Learning discriminative motion cues from a light field ESI is challenging yet crucial for visual obejct tracking. Recent efforts in RGB streams have focused on modeling temporal relations by incorporating an attention-based module. Inspired by this, in this section, we introduce a light field angular-temporal interaction network (ATINet) for object tracking. The overall pipeline is shown in Fig. \ref{framework}. The ATINet includes a dual-stream feature extraction process, where one stream inputs the ESI into the CNN backbone to generate appearance features for the template and search frames separately. The other stream inputs the ESI into our proposed light field angular-temporal modeling framework. The angular-temporal modeling framework consists of a feature extraction baseline and a proposed geometry adaptive selection method. After obtaining the feature maps from both streams, they are each processed through cross-correlation to form their respective correlation feature maps. Finally, we combine the correlation feature maps from both streams and input them into the tracking head. Next, we provide a detailed introduction to the feature extraction baseline and the proposed geometry adaptive selection method.

\subsubsection{Baseline}
\label{Baseline}
The proposed baseline consists of multi-head attention (MHA) blocks and feed-forward networks (FFN). Specifically, we first randomly select two frames to construct a temporal pair $\{ S^{t_1}, S^{t_2}\} \in \mathbb{R}^{H\times W\times 3}$ as input. Secondly, for the ${t_1}$-th frame, $S^{t_1}$ is divided into $N$ non-overlapping patches of resolution $P \times P$, where $N=HW/P^2$.  Thirdly, each of these patches undergoes a separate linear projection and is augmented with learnable position embeddings, resulting in light field patch embeddings $E^{t_1} \in \mathbb{R}^{N\times C}$, where $C$ represents the embedding dimension. Following the same operations, we generate the embeddings for $S^{t_2}$, denoted as $E^{t_2} \in \mathbb{R}^{N\times C}$. Afterward, we concatenate $E^{t_1}$ and $E^{t_2}$ to form a sequence with a length of $2N$, which is then fed into an encoder. In each encoder layer, input features are updated through MHA blocks and FFN. Formally, the process of the $l$-th encoder layer is:
\begin{equation}
\label{baseline1}
[\hat{E}^{t_1}, {\hat{E}^{t_2}}]^{l}= [E^{t_1}, {E^{t_2}}]^{l} + Att([{E}^{t_1}, {{E}^{t_2}}]^{l})
\end{equation}
\begin{equation}
\label{baseline2}
[{E}^{t_1}, {{E}^{t_2}}]^{l+1}= [\hat{E}^{t_1}, {\hat{E}^{t_2}}]^{l} + FFN([\hat{E}^{t_1}, {\hat{E}^{t_2}}]^{l})
\end{equation}
Since the multi-frame of the light field are jointly processed by multi-head self-attention blocks, and attention mechanisms can capture long-term dependency relationships, the modeling of cross-relations and self-relations is seamlessly integrated into each encoder layer. The output of self-attention operation can be defined as:
\begin{equation}
\label{baseline3}
Att([{E}^{t_1}, {{E}^{t_2}}]) = \psi (\frac{[Q_{t_1};Q_{t_2}][K_{t_1};K_{t_2}]^{T}}{\sqrt{d_{k}}})[V_{t_1};V_{t_2}]
\end{equation}
where $\psi$ represents the softmax operation. $Q$, $K$, and $V$ are query, key and value matrices of ${E}^{t_1}$ and ${E}^{t_2}$. For multi-frame embeddings, the attention weights can be expanded to:
\begin{equation}
\label{baseline4}
\frac{[Q_{t_1};Q_{t_2}][K_{t_1};K_{t_2}]^{T}}{\sqrt{d_{k}}} = \frac{[Q_{t_1}K^T_{t_1},Q_{t_1}K^T_{t_2},Q_{t_2}K^T_{t_1},Q_{t_2}K^T_{t_2}]}{\sqrt{d_{k}}}
\end{equation}
Therefore, the self-attention operation can be written as:
\begin{equation}
\label{baseline5}
Att([{E}^{t_1}, {{E}^{t_2}}]) = \psi (\begin{bmatrix}
\varphi ({E}^{t_1}, {{E}^{t_1}}),& \varphi ({E}^{t_1}, {{E}^{t_2}})\\ 
\varphi ({E}^{t_2}, {{E}^{t_1}}),& \varphi ({E}^{t_2}, {{E}^{t_2}})
\end{bmatrix})\begin{bmatrix}
V_{t_1}\\ V_{t_2}
\end{bmatrix}
\end{equation}
where $\varphi(x,y)=Q_xK^T_y/\sqrt{d_{k}}$. In Eq. (\ref{baseline5}), the MHA block explores intra-frame features through self-relations in $\varphi ({E}^{t_1}, {{E}^{t_1}})$ and $\varphi ({E}^{t_2}, {{E}^{t_2}})$, simultaneously modeling inter-frame motion relationships through cross-relations in $\varphi ({E}^{t_1}, {{E}^{t_2}})$ and $\varphi ({E}^{t_2}, {{E}^{t_1}})$. Hence, the motion cues are effectively captured by the MHA block. Finally, the output of the last encoder layer is decoupled for two frames. These features are then reshaped into 2D feature maps based on their original positions, utilized for object tracking.

\subsubsection{Geometry Adaptive Selection}
\begin{figure}[t!]
	\centering
	\includegraphics[width=0.95\linewidth]{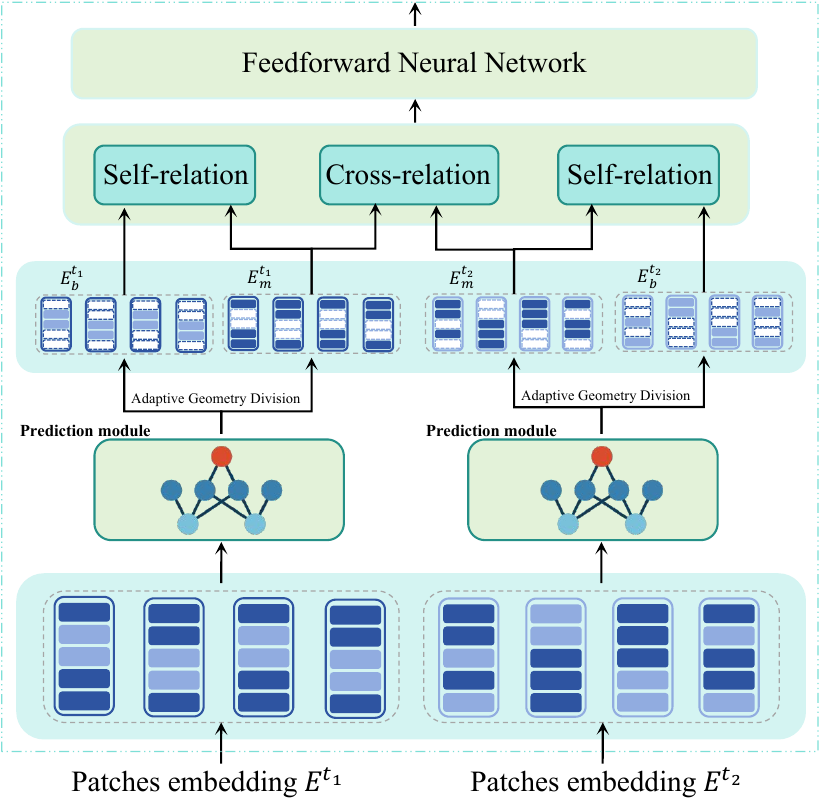}
	\caption{An illustration of the geometry adaptive selection method. It is a plug-and-play module primarily divided into two steps. The first step involves adaptive geometric division using a prediction module. The second step establishes attentions between different embeddings, including self-relation and cross-relation.}
	\label{GAS_framework}
\end{figure}

Introducing the self-attention strategy into the baseline can effectively capture the angular cue correlations between multiple frames in the light field. However, for sparse light field ESI, this direct modeling may diminish feature representation, as certain features from non-geometric structure points can inadvertently aggregate into these representations, weakening the light field’s capability for target-background recognition. To address this issue, we further introduce geometry adaptive selection (GAS) method, which adaptively selects geometric structure features for angular-temporal relation modeling. More specifically, our GAS consists of a lightweight multi-head attention block with an attached trainable prediction module. This module adaptively partitions the ESI embeddings into two groups: $E_m$ and $E_b$, where $E_m$ encapsulates the geometric structure point embeddings crucial for motion perception, while $E_b$ encompasses the remaining non-geometric structure point embeddings. The prediction module can be formulated as:
\begin{equation}
\label{GAS1}
\mathcal{P}=\psi_g(MLP([E^{t_1},E^{t_2}]))
\end{equation}
where MLP is a lightweight multi-layer perceptron and $\psi_g$ is the Gumbel-Softmax. $\mathcal{P} \in \{0,1\}^{2N \times 2}$, which is the one-hot tensor representing the labels of belonging to groups $E_b$ and $E_m$. Next, two columns of zeros are appended to the back of the sequence belonging to frame $t_1$, and two columns of zeros are appended to the front of the sequence belonging to frame $t_2$ in $\mathcal{P}$. We denote the resulting tensors as $\mathcal{\hat{P}} \in \{0,1\}^{2N \times 4}$, which indicate the labels of belonging to groups $E^{t_1}_b$, $E^{t_1}_m$, $E^{t_2}_b$ and $E^{t_2}_m$. Next, we construct a relation matrix using each column element of $\mathcal{\hat{P}}$. The process can be formulated as:
\begin{equation}
\label{GAS2}
\begin{split}
\mathcal{W}_{i,j}=\mathcal{\hat{P}}_{i,0}\mathcal{\hat{P}}_{j,0}+\mathcal{\hat{P}}_{i,1}(\mathcal{\hat{P}}_{j,1}+\mathcal{\hat{P}}_{j,3}) + \\ \mathcal{\hat{P}}_{i,2}\mathcal{\hat{P}}_{j,2} + \mathcal{\hat{P}}_{i,3}(\mathcal{\hat{P}}_{j,3}+\mathcal{\hat{P}}_{j,1})
\end{split}
\end{equation}
where $\mathcal{W} \in \{0,1\}^{2N \times 2N}$, the element $\mathcal{W}_{i,j}$ indicates whether the $i$-th embedding can model relations with the $j$-th embedding and aggregate information. It is noted that $\mathcal{\hat{P}}_{i,0}\mathcal{\hat{P}}_{j,0}$ and $\mathcal{\hat{P}}_{i,2}\mathcal{\hat{P}}_{j,2}$ represent self-relation modeling for the $E_b$ groups of the two frames, without interacting with the $E_m$ groups. $\mathcal{\hat{P}}_{i,1}(\mathcal{\hat{P}}_{j,1}+\mathcal{\hat{P}}_{j,3})$ and $\mathcal{\hat{P}}_{i,3}(\mathcal{\hat{P}}_{j,3}+\mathcal{\hat{P}}_{j,1})$ represent self-relation and cross-relation modeling for the $E_m$ groups of the two frames. Afterward, we calculate the element-wise product of the relation matrix and the attention weight matrix to form the updated attention weight matrix. That is, $\psi_{i,j} \Rightarrow \mathcal{W}_{i,j} \cdot \psi_{i,j}$. Finally, the proposed GAS is integrated into each layer of the baseline. Therefore, Eq. (\ref{baseline2}) can be redefined as:
\begin{equation}
\label{GAS}
\begin{split}
[{E}^{t_1}, {{E}^{t_2}}]^{l+1}= F_{baseline} + GAS([{E}^{t_1}, {{E}^{t_2}}]^{l}) \\
F_{baseline} = [\hat{E}^{t_1}, {\hat{E}^{t_2}}]^{l} + FFN([\hat{E}^{t_1}, {\hat{E}^{t_2}}]^{l})
\end{split}
\end{equation}
It is worth noting that during the training phase, we only trained the GAS module, while the baseline parameters were from a pre-trained Vision Transformer (ViT), which has been proven to be effective in prompt learning \cite{promptSOD}. In this way, the GAS can exclude non-geometric structure points and adaptively choose geometric structure points for angular-temporal relation modeling. This method facilitates the creation of discriminative representations for motion perception via the light field ESI.

\subsection{Training Strategy and Task Expansion}
Although ATINet has established temporal relations among geometric structure points in the light field, training it effectively remains a significant challenge. On one hand, achieving motion awareness in a light field necessitates unsupervised learning capabilities, akin to the human visual system’s ability to establish visual correspondence over time. On the other hand, the limited availability of light field datasets requires GAS to focus more intensely on learning temporal relations. To address this challenge, we propose introducing a self-supervised loss (SSL) as the objective function for the light field. The loss function can be formulated as:
\begin{equation}
\label{SSL}
\mathcal{L}_M = \left \| \mathcal{D}_{\phi }(E\odot \mathcal{M})- E\odot (1-\mathcal{M})\right \|_{2}
\end{equation}
where $\mathcal{M}$ represents a random mask designed to obscure a portion of the embedding features. The $\odot$ symbol indicates element-wise product operation. $E = \mathcal{H}_{GAS}([E^{t_1},E^{t_2}])$ and $\mathcal{H}_{GAS}$ denotes the encoder network with the proposed GAS. $\mathcal{D}_{\phi}$ denotes the lightweight decoder network parameterized by $\phi$, consisting of a small number of MHA blocks and FFN. Additionally, inspired by DropMAE \cite{wu2023dropmae}, we introduce adaptive spatial-attention dropout in $\mathcal{D}_{\phi}$, which constrains interactions among embeddings of geometric structure points within the same frame in the decoder, while simultaneously encouraging more interactions with embeddings of geometric structure points from another frame. In this way, the proposed loss not only enables the training of network in an unsupervised manner but also encourages it to prioritize decoding using inter-frame cues. This compels the baseline with ATINet to focus on object motion for the effective learning of representations supporting temporal matching.

Finally, during the training phase, the total loss is formulated as follows: 
\begin{equation}
\label{MOT1}
\mathcal{L}_{total} = \lambda_1 \mathcal{L}_M + \lambda_2 \mathcal{L}_{cls} + \lambda_3 \mathcal{L}_{reg} 
\end{equation}
where $\lambda_*$ is the weighting factor used to balance the training. The classification branch $\mathcal{L}_{cls}$ is supervised using a Gaussian map generated from the ground truth center and employs focal loss, while the regression branch $\mathcal{L}_{reg}$ is trained with IOU loss.

Furthermore, we develop a light field multiple object tracking framework (AMTrack) based on Trades \cite{Trades}. Since Trades includes a cost volume-based association module to extract re-identification (re-ID) features and further match object similarities across frames, our AMTrack integrates GAS into the re-ID process and establishes a dual-stream re-ID layer to encode both appearance and geometric features. These features are combined for object association during tracking. Consistent with ATINet, the total loss for AMTrack is formulated as follows:
\begin{equation}
\label{VOT1}
\mathcal{L}_{total} = \lambda_1 \mathcal{L}_M + \lambda_2 \mathcal{L}_{det} + \lambda_3 \mathcal{L}_{CVA} 
\end{equation}
where $\lambda_*$ is the weighting factor used to balance the training. The $\mathcal{L}_{det}$ represents the object detection loss. The CVA loss $\mathcal{L}_{CVA}$ is supervised to learn an effective re-ID appearance embedding.

\section{Experiments}
\subsection{R8LUT Datasets}
Currently available light field datasets are based on static images and lack light field video data in real low-light scenarios. Therefore, we manually collect light field videos in low-light environments using the advanced Raytrix R8 camera\cite{Raytrix}. The videos feature a diverse array of objects including glass spheres, toy cars, industrial nuts, and fish. Each scene is carefully composed of multiple, similar objects, creating a dynamic and complex motion flow within the video stream. Subsequently, we record these scenes using RxLive at a frame rate of 25 frames per second. RxLive captures this data in a raw format as rays, which are then processed using the SDK interface of the Raytrix camera. Each light field frame undergoes parsing into multi-view images, which are further constructed into a light field ESI representation, as illustrated in Fig. 3. The spatial resolution of our light field is $1080 \times 1920$, coupled with an angular resolution of $5 \times 5$.

With the setup described above, we initially acquire over 12,500 light fields. Subsequently, we inspect the quality of each light field and discard samples with discrepancies. Following the filtering process, we retain 3,500 high-quality light fields, which ultimately form 26 light field video sequences, each containing approximately 150 frames. Next, we engage 30 participants to identify moving objects within each light field sequence. In accordance with the annotation standards of MOT16 \cite{2016motchallenge}, we assign a unique numerical identifier to each moving object within every video in the light field video series. Considering that each video contains 7 to 8 moving objects, this process results in the creation of the light field MOT dataset. Furthermore, we compile 160 objects exhibiting various movements from the light field MOT dataset and incorporate datasets of moving objects from other studies. This comprehensive effort results in the formation of the light field SOT dataset.

Finally, following the protocol established by MOT16, we divided the generated light field MOT datasets into training and testing subsets. The training subset contains 26 videos with 1,800 samples, and the testing subset includes an equivalent number of videos, totaling 1,500 samples. For the light field SOT dataset, we distributed the datasets into training and testing sets at a 6:4 ratio. The training set comprises 102 videos with 3,500 samples, whereas the testing set includes 71 videos, amassing 15,380 samples. The proposed dataset is available at \url{https://github.com/lfvision/LightFieldTracking}.

\subsubsection{Metrics}
To analyze the results of various methods in the SOT task, we employ three widely used evaluation metrics for quantitative performance assessment: precision, success, and normalized precision \cite{OTB}. Additionally, we generate precision, normalized precision, and success plots for evaluation. For the MOT task, we utilize four evaluation metrics for performance assessment: false positives (FP) \cite{1evaluationMOT}, false negatives (FN) \cite{1evaluationMOT}, ID switches (IDS) \cite{1evaluationMOT}, IDF1 \cite{3evaluationID}, and Multiple Object Tracking Accuracy (MOTA) \cite{1evaluationMOT}. 
\subsubsection{Implementation Details}
We employ the PyTorch toolbox for our light field SOT and MOT tasks on a system equipped with four NVIDIA RTX 3090 GPUs. In the ATINet, our baseline is parameterized using a pre-trained ViT. The GAS module is integrated across all blocks of the baseline with settings of 12 layers depth, 16 attention heads, and an embedding dimension of 768. The mask rate in ALS is set at 0.5, and the decoder depth is set at 4. We optimize ATINet with a batch size of 96 for 25 epochs. For AMTrack, training spans 70 epochs with a maximum frame distance of 10, selecting 2 frames in total, and a training image size of $544 \times 960$. During tracking, the detection and tracking thresholds are set at 0.5 and 0.4, respectively.

\subsection{Comparisons with State-of-the-arts Methods}
\subsubsection{Evaluation for Single Object Tracking}

\begin{table}[!t]
\centering
\caption{Comparison on the proposed light field tracking dataset with hybrid tracking framework. The best results are marked in bold.}
\label{tab:tracker_two_stream_siamese}
\resizebox{\linewidth}{!}{
	\begin{tabular}{ccccc}
		\toprule[1.3pt]
	Trackers     & Years & Success$\uparrow$   & Precision$\uparrow$   & Norm.Prec.$\uparrow$  \\
		\midrule 
	SiamRPN \cite{siamrpn}           & CVPR18 & 0.38 & 0.46 & 0.50\\
	  SiamRPN++ \cite{siamrpn++}       & CVPR19 & 0.57 & 0.75 & 0.76\\
	  SiamDW \cite{SiamDW}             & CVPR19 & 0.44 & 0.55 & 0.58\\
        ATOM \cite{ATOM}                 & CVPR19 & 0.42 & 0.49 & 0.50\\
        DiMP \cite{DiMP}                 & ICCV19 & 0.59 & 0.71 & 0.72\\
        SiamFC++ \cite{xu2020siamfc++}   & AAAI20 & 0.57 & 0.74 & 0.76\\
        PrDiMP \cite{PrDiMP}             & CVPR20 & 0.57 & 0.68 & 0.69\\
        SiamCAR \cite{siamcar}           & CVPR21 & 0.55 & 0.74 & 0.76\\
        SiamAPN \cite{siamAPN}           & ICRA21 & 0.45 & 0.65 & 0.68\\
        STMTrack \cite{stmtrack}         & CVPR21 & 0.55 & 0.70 & 0.71\\
        LFtrack \cite{wang2022visual}    & TII23  & 0.36 & 0.46 & 0.51\\
        ToMP \cite{ToMP}                 & CVPR20 & 0.60 & 0.72 & 0.72\\
	  TrTr \cite{trtr}                 & ArXiv21 & 0.54 & 0.67 & 0.69\\
   	TaMOs \cite{TaMOs}               & WACV24 & 0.57 & 0.69 & 0.70\\
        MCITrack \cite{MCITrack}         & AAAI25 & 0.55 & 0.66 & 0.67\\
		\midrule
        STARK \cite{STARK}                      & ICCV21 & 0.57 & 0.67 & 0.68\\
	  OSTrack \cite{OSTrack}                  & ECCV22 & 0.59 & 0.71 & 0.71\\
	  SimTrack \cite{simtrack}                & ECCV22 & 0.61 & 0.74 & 0.75\\
	  MixFormerV2 \cite{cui2024mixformerv2}   & NeurIPS23 & 0.53 & 0.64 & 0.66\\
	  ZoomTrack \cite{kou2024zoomtrack}       & NeurIPS23 & 0.61 & 0.73 & 0.74\\
	  GRM \cite{GRM}                          & CVPR23 & 0.63 & 0.75 & 0.76\\
   	DropTrack \cite{wu2023dropmae}          & CVPR23 & 0.62 & 0.74 & 0.74\\
	  ODTrack \cite{odtrack}                  & AAAI24 & 0.56 & 0.67 & 0.68\\
      SUTrack \cite{Sutrack}                  & AAAI25 & 0.60 & 0.71 & 0.73\\
		\midrule
        \rowcolor[HTML]{e6e6e6}
        {\textbf{ATINet}}         &-   & \textbf{0.64}	& \textbf{0.79}	& \textbf{0.81}\\
		\bottomrule[1.3pt]
	\end{tabular}
    }
\end{table}

\begin{figure*}[t!]
	\centering
	\includegraphics[width=0.95\linewidth]{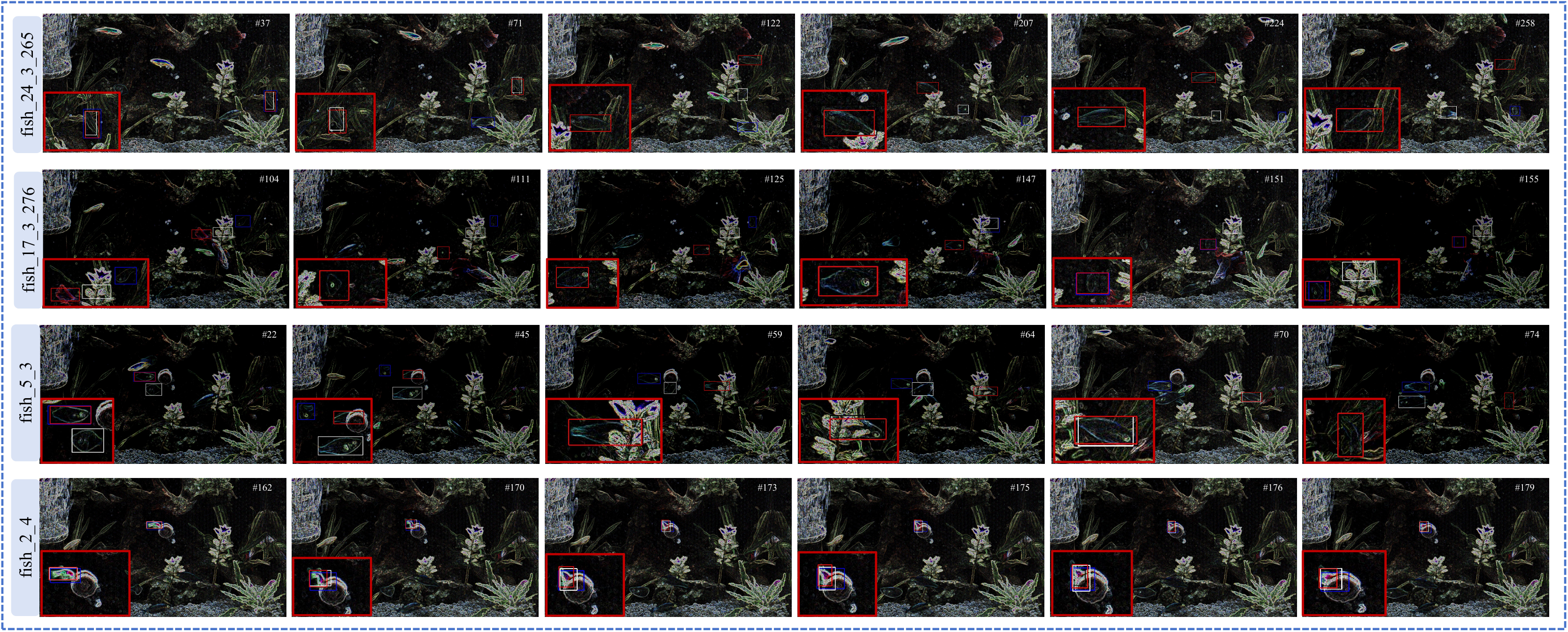}
	\caption{Visual comparison between our proposed tracker and other methods in the low-light scenes. The black box indicates the ground truth, the red box represents our tracker, the blue box corresponds to DropTrack, and the white box denotes OSTrack.}
	\label{Figure5_SOT}
\end{figure*}

As the state-of-the-art in the light field ESI domain remains largely unexplored, we conduct a comprehensive evaluation of our proposed method by comparing it with 24 existing state-of-the-art methods on light field ESI datasets. Specifically, our competitors include 15 hybrid methods: ATOM \cite{ATOM}, DiMP \cite{DiMP}, PrDiMP \cite{PrDiMP}, SiamRPN \cite{siamrpn}, SiamRPN++ \cite{siamrpn++}, SiamDW \cite{SiamDW}, SiamFC++ \cite{xu2020siamfc++}, SiamCAR \cite{siamcar}, STMTrack \cite{stmtrack}, SiamAPN \cite{siamAPN}, TrTr \cite{trtr}, ToMP \cite{ToMP}, TaMOs \cite{TaMOs}, MCITrack \cite{MCITrack}, and LFtrack \cite{wang2022visual}; and 9 pure transformer methods: STARK \cite{STARK}, OSTrack \cite{OSTrack}, SimTrack \cite{simtrack},  DropTrack \cite{wu2023dropmae}, GRM \cite{GRM}, MixFormerV2 \cite{cui2024mixformerv2}, ZoomTrack \cite{kou2024zoomtrack}, SUTrack \cite{Sutrack}, and ODTrack \cite{odtrack}. Notably, LFtrack also employs light field-based tracking. For a fair comparison, we re-train these trackers on our training dataset using their default parameter settings, rather than directly testing them on the testing subset.

The quantitative results are listed in Table \ref{tab:tracker_two_stream_siamese}, where we observe that our method outperforms the other methods, with higher success and precision. This is primarily because trackers are generally optimized for RGB image tracking and struggle with processing and modeling features from light field ESI inputs, particularly in low-ligtht scenes. Furthermore, these trackers tend to focus predominantly on spatial dimension information extraction, neglecting the angular-temporal dimension. This oversight leads to their inability to leverage visual motion cues within light field videos effectively. 
Owing to the specialized design of our GAS module tailored for light field ESI, its integration into the tracking framework significantly improves performance over existing methods. Furthermore, by capturing geometric structural cues across angular-temporal dimensions, the proposed approach effectively extracts light field motion features, enabling accurate target localization in low-light scenes.

To observe the visual quality of the tracking results produced by our method and others, we present several representative examples in Fig. \ref{Figure5_SOT}. Although recent RGB imaging-based methods have made significant progress, they still fall short in exploiting the geometric structural information of light fields to accurately locate moving targets in low-light scenes, particularly those involving targets like fish that exhibit significant deformations, similarity interference, and motion blur. In contrast, our method successfully distinguishes moving objects in these complex scenes, demonstrating the superiority of our proposed method in light field tracking.

\subsubsection{Evaluation for Multiple Object Tracking}

\begin{table}[t!]
\centering
\caption{Quantitative Comparison of the State-of-the-art Methods on the light field MOT tracking dataset.}
\label{tab:MOTtracker}
\resizebox{\linewidth}{!}{
\begin{tabular}{ccccccc}
\toprule[1.3pt]
 Model   &Years      & MOTA$\uparrow$  & IDF1$\uparrow$  & IDS$\downarrow$  & FP$\downarrow$    & FN$\downarrow$     \\ 
\midrule
CenterTrack \cite{CenterTrack} &ECCV20 & 87.1  & 71.2  & 339 & 626  & \textbf{2029}  \\
Trades \cite{Trades} &CVPR21      & 87.3  & 83.3  & 123 & 416  & 2394  \\
ByteTrack \cite{ByteTrack} &ECCV22   & 74.2  & 81.2  & \textbf{66} & 2770  & 3140\\
Hybrid-SORT \cite{Hybrid-SORT} &AAAI24 & 85.3  & 75.3  & 201 & \textbf{310}  & 2904 \\
SparseTrack \cite{SparseTrack} &TCSVT25 & 86.4  & 83.8  & 130 & 418  & 2600 \\
\midrule
\rowcolor[HTML]{e6e6e6}
\textbf{AMTrack} & -    & \textbf{87.5}   & \textbf{85.4}    & 125  & 455  & 2306 \\
\bottomrule[1.3pt]
\end{tabular}
}
\end{table}

\begin{figure*}[t!]
	\centering
	\includegraphics[width=\linewidth]{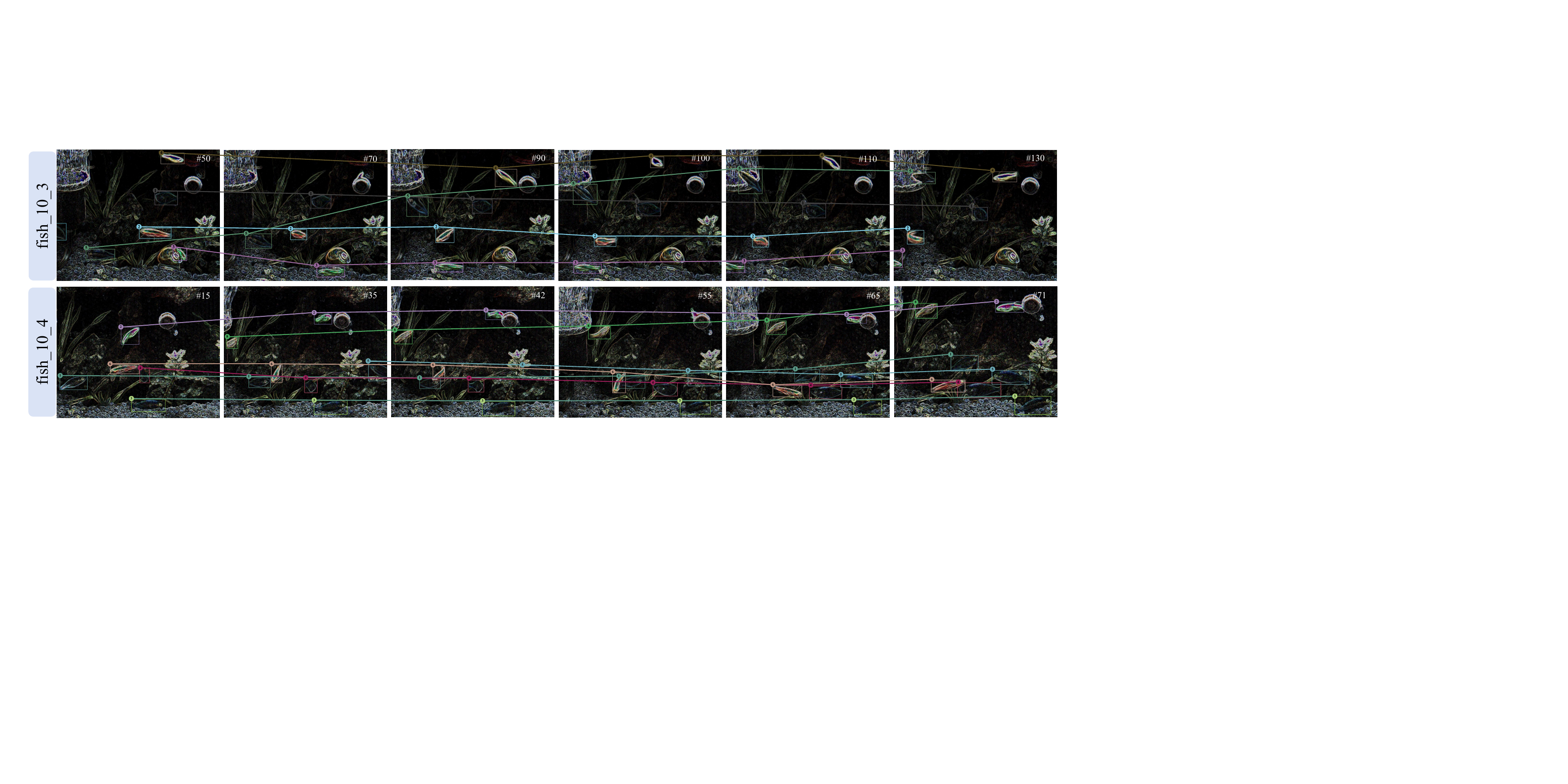}
	\caption{Visual tracking results of the proposed light field MOT algorithm on some scenes.}
	\label{Figure6_MOT}
\end{figure*}

To fully evaluate the performance of our method in the light field MOT task, we compared our proposed method with four other state-of-the-art MOT methods that use light field ESI as input. These methods include CenterTrack \cite{CenterTrack}, Trades \cite{Trades}, ByteTrack \cite{ByteTrack}, and Hybrid-SORT \cite{Hybrid-SORT}. To ensure a fair comparison, we retrained these trackers using their default parameter settings on our training dataset.

Table \ref{tab:MOTtracker} presents a comparison between our proposed MOT method and other algorithms. As indicated in Table \ref{tab:MOTtracker}, our method achieved the highest MOTA score among these state-of-the-art methods. This underscores the superior overall performance of our proposed method compared to competing algorithms. Additionally, our method also recorded the highest IDF1 score among all methods compared, demonstrating its exceptional capability in maintaining long-term tracking of consistent target trajectories. Moreover, by incorporating temporal motion information from the light field, our algorithm increasingly concentrates on time-series-related objects. This enhancement allows the tracker to accurately match most objects, which exhibit comparatively low miss detection rates.

Fig. \ref{Figure6_MOT} illustrates several frames of tracking results generated by our method. The consistency of the estimated trajectories is indicated by bounding boxes marked with identical colors and ID numbers. From the figure, it is evident that our method can successfully locate and identify most objects within complex low-light environments. These results not only confirm that the light field ESI effectively captures complex object information for MOT tasks, but also highlight that our proposed method is adept at accurately learning re-identification embedding information crucial for tracking associations.

\subsection{Ablation Study and Analysis}
\subsubsection{Effectiveness of Key Components}
\begin{table}[t!]
\centering
\caption{Ablation study for Key Components.}
\label{tab:Components}
\resizebox{\linewidth}{!}{
\begin{tabular}{ccccccc}
\toprule[1.3pt]
 & ESI   & GAS & SSL & Success $\uparrow$  & Precision$\uparrow$  & Norm.Prec.$\uparrow$  \\ 
\midrule
1& & &  & 0.58 & 0.74 & 0.76 \\
2& $\checkmark$& &  & 0.59 & 0.73 & 0.74 \\
3& $\checkmark$&$\checkmark$ &  & 0.62 & 0.77 & 0.78 \\
4& $\checkmark$& &$\checkmark$ & 0.60 & 0.75 & 0.77 \\
\rowcolor[HTML]{e6e6e6}
5& $ \checkmark$&$\checkmark$ &$\checkmark$  & \textbf{0.64} & \textbf{0.79} & \textbf{0.81} \\
\bottomrule[1.3pt]
\end{tabular}
}
\end{table}

In this subsection, we conduct ablation studies to assess the significance of various components. The performance of various methods is displayed in Table \ref{tab:Components}. The first row illustrates the baseline results, which do not incorporate light field components and instead use the single view of the light field as the direct input. This row shows lower performance metrics, suggesting that basic spatial cues alone are inadequate for fully capturing the motion states of objects in complex low-light scenes. In the second row, we incorporate the proposed light field ESI as input. However, we observe no significant enhancement in tracker performance. This is mainly due to traditional feature modeling methods being ineffective at extracting the sparse structured cues inherent in ESI. In contrast, the third row introduces GAS, which specifically focuses on light field angular-temporal modeling, allowing the tracker to better comprehend complex scenes. Meanwhile, in the fourth row, our proposed SSL significantly enhances the model's training efficiency on limited datasets. Finally, in the fifth row, by integrating all three proposed light field components, the tracker achieves an optimal configuration, thereby validating the effectiveness of our approach.

\subsubsection{Analysis on Different ESI Representation}
\begin{figure}[t!]
	\centering
	\includegraphics[width=\linewidth]{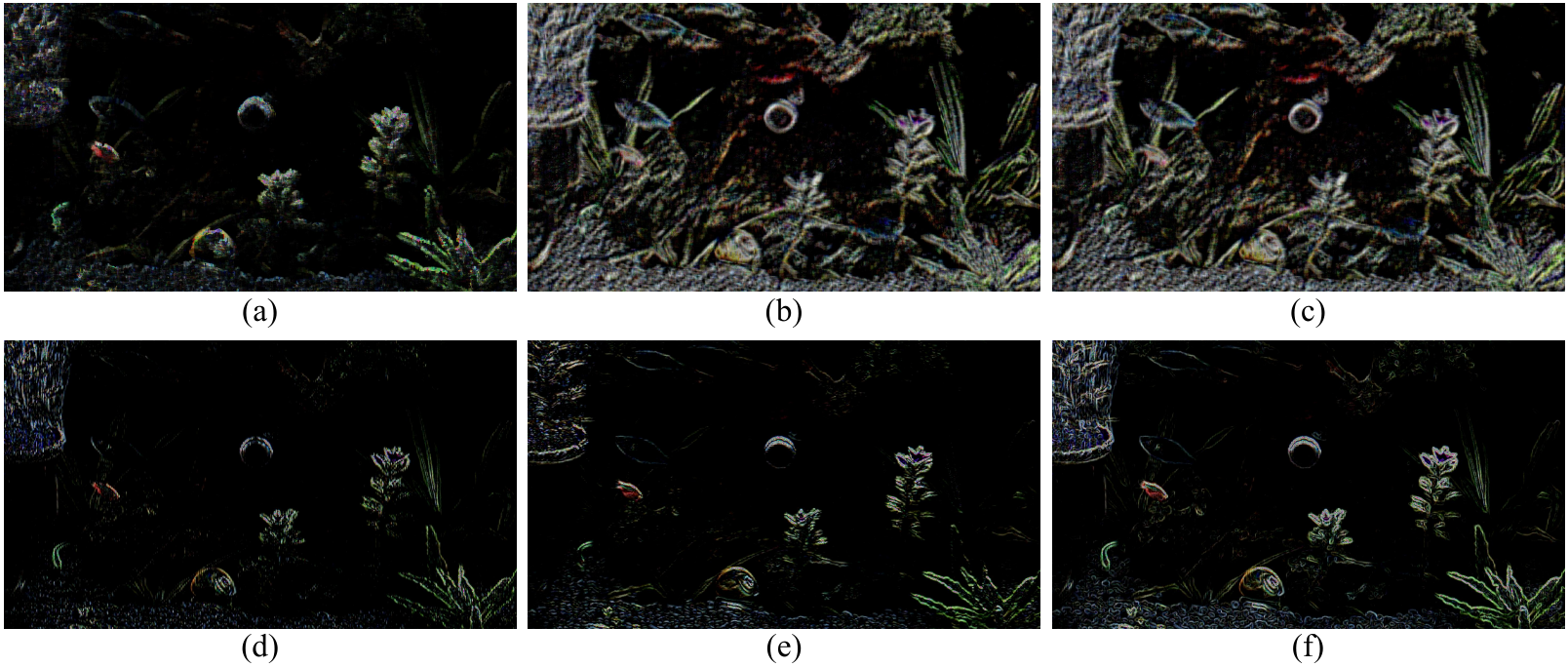}
	\caption{Visual results of different light field representations. (a) $ESI_{max}$. (b) $ESI_{mean}$. (c) $ESI_{sum}$. (d) $ESI_{H}$. (e) $ESI_{V}$. (f) $ESI$.}
	\label{Figure8_ESI}
\end{figure}

In our proposed ESI, we select the gradient values from the central view ($u=U/2$) and calculate the amplitude of the horizontal and vertical EPI gradient images to form the final light field ESI representation. In this subsection, we explore various light field representations by adopting alternative settings. Specifically, we capture the maximum, mean, and sum values across the $u$ and $v$ dimensions from the horizontal and vertical EPI gradient images, defining these as $ESI_{max}$, $ESI_{mean}$, and $ESI_{sum}$. Additionally, we directly calculate the amplitude of the horizontal and vertical EPI gradient images to establish directional representations from a single viewpoint, designated as $ESI_{H}$ and $ESI_{V}$. As shown in Table \ref{tab:ESI}, the metrics confirm that the ESI representation significantly improves performance. The statistically derived $ESI_{mean}$ and $ESI_{sum}$ representations struggle with continuity in edge regions, as depicted in Fig. \ref{Figure8_ESI}. Moreover, the unidirectional $ESI_{H}$ and $ESI_{V}$ representations fail to capture significant changes in light field targets. Conversely, the proposed ESI can capture the geometric structure and motion of the target across both horizontal and vertical directions, thereby achieving optimal results.

\begin{table}[t!]
\centering
\caption{Ablation study for Different ESI Representation.}
\label{tab:ESI}
\resizebox{\linewidth}{!}{
\begin{tabular}{ccccccc}
\toprule[1.3pt]
 & Input   & Success$\uparrow$  & Precision$\uparrow$  & Norm.Prec.$\uparrow$ \\ 
\midrule
1&ESI$_{Max}$  & 0.56 & 0.70 & 0.71 \\
3&ESI$_{Mean}$ & 0.56 & 0.71 & 0.73 \\
4&ESI$_{Sum}$  & 0.57 & 0.73 & 0.74 \\
5&ESI$_V$      & 0.59 & 0.74 & 0.75 \\
6&ESI$_H$      & 0.58 & 0.73 & 0.75 \\
\rowcolor[HTML]{e6e6e6}
7&\textbf{ESI}          & \textbf{0.64} & \textbf{0.79} & \textbf{0.81} \\
\bottomrule[1.3pt]
\end{tabular}
}
\end{table}

\subsubsection{Analysis on Different GAS Setting}
\begin{table}[t!]
\centering
\caption{Ablation Study for Different Settings with the Proposed GAS.}
\label{tab:GAS}
\resizebox{\linewidth}{!}{
\begin{tabular}{cccccc}
\toprule[1.3pt]
 & Type   & Success$\uparrow$  & Precision$\uparrow$  & Norm.Prec.$\uparrow$ \\ 
\midrule
1&baseline  & 0.60 & 0.75 & 0.77 \\
2&GAS$_{shallow}$      & 0.63 & \textbf{0.79} & 0.80 \\
3&GAS$_{depth}$      & 0.62 & 0.78 & 0.78 \\
4&GAS$_{intra}$      & 0.59 & 0.74 & 0.76 \\
5&GAS$_{inter}$      & 0.61 & 0.76 & 0.78 \\
\rowcolor[HTML]{e6e6e6}
6&\textbf{GAS}          & \textbf{0.64} & \textbf{0.79} & \textbf{0.81} \\
\bottomrule[1.3pt]
\end{tabular}
}
\end{table}

In the proposed GAS, GAS is embedded into all layer of neurons in the baseline model. In this subsection, we assess the effectiveness of GAS in light field angular-temporal modeling by selectively removing GAS branches from the deep and shallow layers of our method. Additionally, we modify the configuration of the relation matrix within the GAS branches to evaluate the efficacy of our relational modeling. The results are presented in Table 5. Specifically, $GAS_{depth}$ and $GAS_{shallow}$ refer to the integration of GAS into the deep and shallow layers of the baseline, respectively. $GAS_{intra}$ and $GAS_{inter}$ represent the modeling of only intra-frame and inter-frame correlations, respectively, within the $E_m$ groups for two frames. Compared to $GAS_{depth}$ and $GAS_{shallow}$, our method shows significant improvements. The comparison results confirm the effectiveness of our proposed GAS module in the both deep and shallow layers. Additionally, our method outperforms both $GAS_{intra}$ and $GAS_{inter}$. This indicates that angular-temporal modeling for light fields must consider not only the temporal correlations across frames but also the intra-frame structural correlations.

\subsubsection{Analysis of Computational Complexity}

\begin{table}[t!]
\centering
\caption{The Computational Complexity Analysis in the Networks With or Without the Proposed GAS.}
\label{tab:Params}
\resizebox{\linewidth}{!}{
\begin{tabular}{cccc}
\toprule[1.3pt]
Type   & FLOPs$\downarrow$   & Params$\downarrow$   & Time$\downarrow$  \\ 
\midrule
baseline  & 39.07 & 99.78 & 0.019 $\pm$ 0.0001 \\
baseline+SSL     & 39.44 & 100.77 & 0.053 $\pm$ 0.0003 \\
\rowcolor[HTML]{e6e6e6}
\textbf{baseline+GAS+SSL (Ours)}      & 44.35 & 113.31 & 0.066 $\pm$ 0.0010 \\
\bottomrule[1.3pt]
\end{tabular}
}
\end{table}
The main computational demand of our light field ATINet arises from the prediction module in GAS and the calculation of relationship matrices, as well as the decoder part in SSL. To assess the complexity of our method, we compare the computational complexity of the network embedded with  GAS and SSL to that of the original network. Results in Table \ref{tab:Params} showcase the parameters, FLOPs, and inference time for networks with and without GAS and SSL. It is evident from Table \ref{tab:Params} that, our SSL does not incur additional parameter costs due to its lightweight decoding design. Furthermore, as a plug-and-play module, our GAS introduces only a minimal computational load to the original network. Consequently, when our method is integrated into downstream tasks, there is merely a slight increase in computational complexity. 
\begin{figure}[t!]
	\centering
	\includegraphics[width=\linewidth]{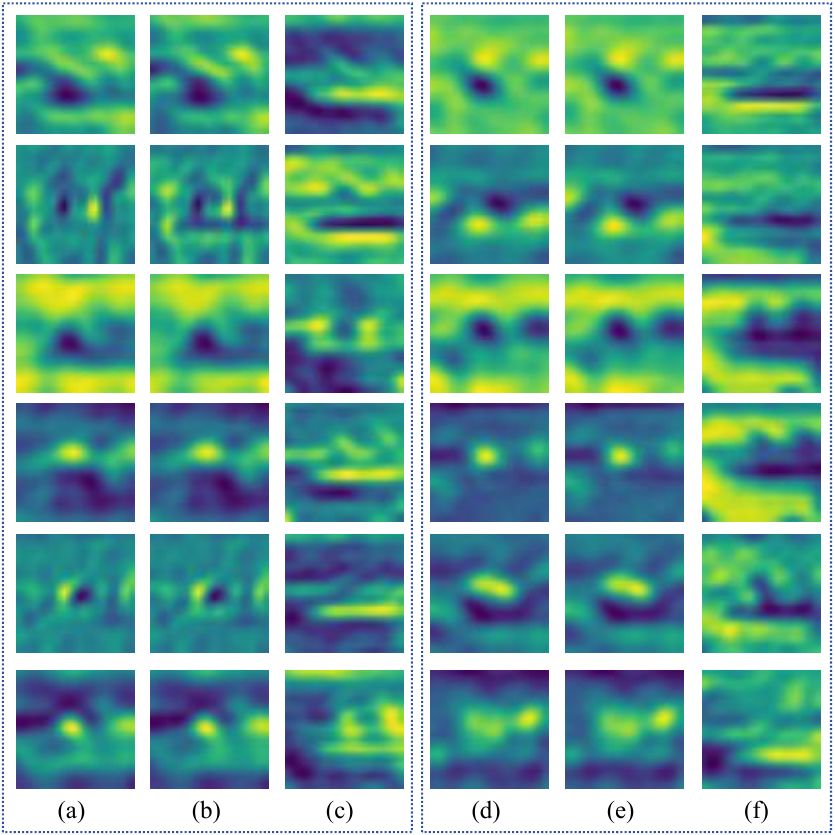}
	\caption{Visualization of feature maps following Siamese network. (a) and (d) Feature maps generated by our method, which combine correlation feature maps from both streams. (b) and (e) Feature maps produced with our proposed method with GAS deployment. (c) and (f) Feature maps produced without deploying GAS.}
	\label{Figure_map}
\end{figure}

\begin{figure}[t!]
	\centering
	\includegraphics[width=\linewidth]{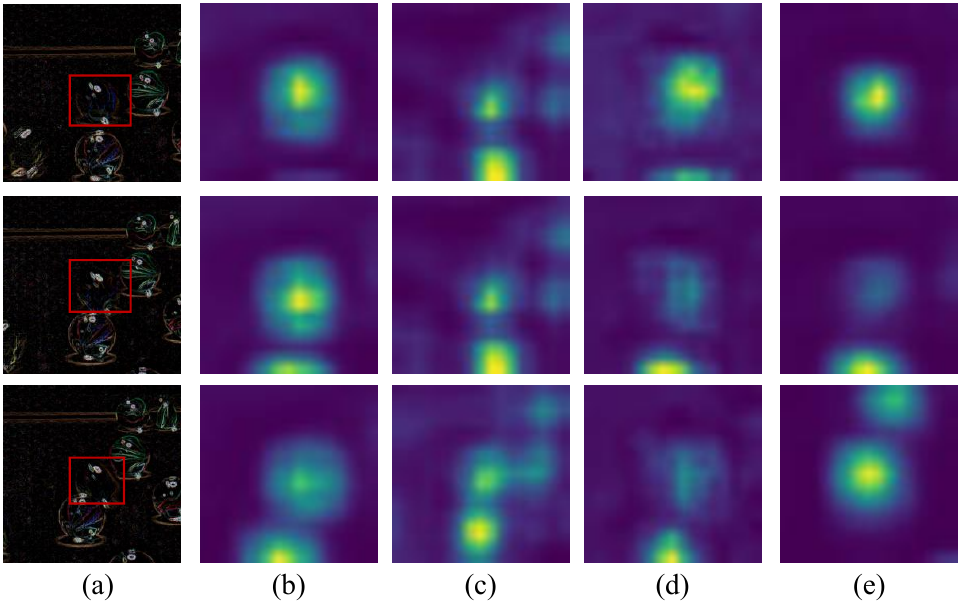}
	\caption{Visualization of tracking score maps across various frames. (a) Search frame within the tracking sequence. (b) Results generated by our method. (c) Results obtained without deploying SSL. (d) Results obtained without deploying GAS. (e) Results achieved without using light field ESI.}
	\label{Figure_score}
\end{figure}
\subsubsection{Visualization Analysis}
In the ATINet framework, the light field ESI input is processed by both the backbone and the proposed method to generate their respective correlation feature maps. These maps from both streams are then directly combined for prediction. In this subsection, we visualize the feature maps from both streams and the final composite feature map to assess the effectiveness of our method. As illustrated in Fig. \ref{Figure_map}, the feature maps produced by our proposed method are notably clearer than those from the backbone, enabling preliminary identification of the target's location. This suggests that the conventional backbone struggles to extract meaningful light field information from the ESI, whereas our proposed method successfully generates discriminative light field features. Furthermore, we visualize the tracking score maps across different frames. The results, presented in Fig. \ref{Figure_score}, demonstrate that our method effectively produces accurate tracking score maps, even in complex low-light scenes, validating the effectiveness of our proposed light field object tracking framework.

\section{Conclusion}
In this paper, we observe that the geometric structure within the light field is effectively represented by angular variations in the angular domain. Inspired by this observation, we propose a succinct light field ESI representation. This representation aims to reduce redundancy in high-dimensional light fields and enhance the expression of visual cues in low-light scenes. To learn angular-aware representations from the geometric structure and angular-temporal interaction cues of the light field, especially in complex low-light scenes, we introduce ATINet for light field object tracking, which includes multiple GAS modules and an SSL mechanism. The GAS module selectively gathers sparse geometric structure cues across angular-temporal dimensions and models the correlations within and between frames in the angular domain. The SSL optimizes light field angular-temporal learning through a self-supervised manner, further enhancing the interaction of geometric features across the temporal domain. Additionally, we establish a large-scale light field video benchmark, providing significant data support for light field SOT and MOT tasks. Our experiments validate the effectiveness of our proposed methods and report state-of-the-art results on these two representative visual tasks. Furthermore, through analysis and comparison of various modules, the experiments show that our proposed method is feasible and effective for achieving target-level temporal matching tasks in low-light scenes.

\appendices

\ifCLASSOPTIONcaptionsoff
  \newpage
\fi

\bibliographystyle{IEEEtran}
\bibliography{BIB_TIP}

@article{TPAMIdisparityestimation,
  title={Disentangling light fields for super-resolution and disparity estimation},
  author={Wang, Yingqian and Wang, Longguang and Wu, Gaochang and Yang, Jungang and An, Wei and Yu, Jingyi and Guo, Yulan},
  journal={IEEE Transactions on Pattern Analysis and Machine Intelligence},
  volume={45},
  number={1},
  pages={425--443},
  year={2022},
  publisher={IEEE}
}

@article{TPAMIreconstruction,
  title={Light field reconstruction via deep adaptive fusion of hybrid lenses},
  author={Jin, Jing and Guo, Mantang and Hou, Junhui and Liu, Hui and Xiong, Hongkai},
  journal={IEEE Transactions on Pattern Analysis and Machine Intelligence},
  year={2023},
  publisher={IEEE}
}

@article{chen2022deep,
  title={Deep light field spatial super-resolution using heterogeneous imaging},
  author={Chen, Yeyao and Jiang, Gangyi and Yu, Mei and Xu, Haiyong and Ho, Yo-Sung},
  journal={IEEE Transactions on Visualization and Computer Graphics},
  year={2022},
  publisher={IEEE}
}

@inproceedings{CenterTrack ,
  title={Tracking objects as points},
  author={Zhou, Xingyi and Koltun, Vladlen and Kr{\"a}henb{\"u}hl, Philipp},
  booktitle={European conference on computer vision},
  pages={474--490},
  year={2020},
  organization={Springer}
}

@inproceedings{Trades,
  title={Track to detect and segment: An online multi-object tracker},
  author={Wu, Jialian and Cao, Jiale and Song, Liangchen and Wang, Yu and Yang, Ming and Yuan, Junsong},
  booktitle={Proceedings of the IEEE/CVF conference on computer vision and pattern recognition},
  pages={12352--12361},
  year={2021}
}

@inproceedings{ByteTrack,
  title={Bytetrack: Multi-object tracking by associating every detection box},
  author={Zhang, Yifu and Sun, Peize and Jiang, Yi and Yu, Dongdong and Weng, Fucheng and Yuan, Zehuan and Luo, Ping and Liu, Wenyu and Wang, Xinggang},
  booktitle={European conference on computer vision},
  pages={1--21},
  year={2022},
  organization={Springer}
}

@inproceedings{Hybrid-SORT,
  title={Hybrid-sort: Weak cues matter for online multi-object tracking},
  author={Yang, Mingzhan and Han, Guangxin and Yan, Bin and Zhang, Wenhua and Qi, Jinqing and Lu, Huchuan and Wang, Dong},
  booktitle={Proceedings of the AAAI Conference on Artificial Intelligence},
  volume={38},
  number={7},
  pages={6504--6512},
  year={2024}
}

@article{MAC,
  title={Light field saliency detection with deep convolutional networks},
  author={Zhang, Jun and Liu, Yamei and Zhang, Shengping and Poppe, Ronald and Wang, Meng},
  journal={IEEE Transactions on Image Processing},
  volume={29},
  pages={4421--4434},
  year={2020},
  publisher={IEEE}
}

@misc{Raytrix,
title = {Raytrix GmbH.},
howpublished = {\url{https://raytrix.de/. Accessed: 2019-02-03}}
}

@article{2016motchallenge,
  title={MOT16: A benchmark for multi-object tracking},
  author={Milan, Anton and Leal-Taix{\'e}, Laura and Reid, Ian and Roth, Stefan and Schindler, Konrad},
  journal={arXiv preprint arXiv:1603.00831},
  year={2016}
}

@article{1evaluationMOT,
  title={Evaluating multiple object tracking performance: the clear mot metrics},
  author={Bernardin, Keni and Stiefelhagen, Rainer},
  journal={EURASIP Journal on Image and Video Processing},
  volume={2008},
  pages={1--10},
  year={2008},
  publisher={Springer}
}

@inproceedings{3evaluationID,
  title={Performance measures and a data set for multi-target, multi-camera tracking},
  author={Ristani, Ergys and Solera, Francesco and Zou, Roger and Cucchiara, Rita and Tomasi, Carlo},
  booktitle={European conference on computer vision},
  pages={17--35},
  year={2016},
  organization={Springer}
}

@inproceedings{ToMP,
  title={Transforming model prediction for tracking},
  author={Mayer, Christoph and Danelljan, Martin and Bhat, Goutam and Paul, Matthieu and Paudel, Danda Pani and Yu, Fisher and Van Gool, Luc},
  booktitle={Proceedings of the IEEE/CVF conference on computer vision and pattern recognition},
  pages={8731--8740},
  year={2022}
}

@inproceedings{xu2020siamfc++,
  title={Siamfc++: Towards robust and accurate visual tracking with target estimation guidelines},
  author={Xu, Yinda and Wang, Zeyu and Li, Zuoxin and Yuan, Ye and Yu, Gang},
  booktitle={Proceedings of the AAAI conference on artificial intelligence},
  volume={34},
  number={07},
  pages={12549--12556},
  year={2020}
}

@inproceedings{siamrpn,
  title={High performance visual tracking with siamese region proposal network},
  author={Li, Bo and Yan, Junjie and Wu, Wei and Zhu, Zheng and Hu, Xiaolin},
  booktitle={Proceedings of the IEEE conference on computer vision and pattern recognition},
  pages={8971--8980},
  year={2018}
}

@inproceedings{siamrpn++,
  title={Siamrpn++: Evolution of siamese visual tracking with very deep networks},
  author={Li, Bo and Wu, Wei and Wang, Qiang and Zhang, Fangyi and Xing, Junliang and Yan, Junjie},
  booktitle={Proceedings of the IEEE/CVF conference on computer vision and pattern recognition},
  pages={4282--4291},
  year={2019}
}

@inproceedings{siamcar,
  title={SiamCAR: Siamese fully convolutional classification and regression for visual tracking},
  author={Guo, Dongyan and Wang, Jun and Cui, Ying and Wang, Zhenhua and Chen, Shengyong},
  booktitle={Proceedings of the IEEE/CVF conference on computer vision and pattern recognition},
  pages={6269--6277},
  year={2020}
}

@inproceedings{stmtrack,
  title={Stmtrack: Template-free visual tracking with space-time memory networks},
  author={Fu, Zhihong and Liu, Qingjie and Fu, Zehua and Wang, Yunhong},
  booktitle={Proceedings of the IEEE/CVF conference on computer vision and pattern recognition},
  pages={13774--13783},
  year={2021}
}

@inproceedings{siamAPN,
  title={Siamese anchor proposal network for high-speed aerial tracking},
  author={Fu, Changhong and Cao, Ziang and Li, Yiming and Ye, Junjie and Feng, Chen},
  booktitle={2021 IEEE International Conference on Robotics and Automation (ICRA)},
  pages={510--516},
  year={2021},
  organization={IEEE}
}

@article{trtr,
  title={Trtr: Visual tracking with transformer},
  author={Zhao, Moju and Okada, Kei and Inaba, Masayuki},
  journal={arXiv preprint arXiv:2105.03817},
  year={2021}
}

@inproceedings{simtrack,
  title={Backbone is all your need: A simplified architecture for visual object tracking},
  author={Chen, Boyu and Li, Peixia and Bai, Lei and Qiao, Lei and Shen, Qiuhong and Li, Bo and Gan, Weihao and Wu, Wei and Ouyang, Wanli},
  booktitle={European Conference on Computer Vision},
  pages={375--392},
  year={2022},
  organization={Springer}
}

@inproceedings{ARTrack,
  title={Autoregressive visual tracking},
  author={Wei, Xing and Bai, Yifan and Zheng, Yongchao and Shi, Dahu and Gong, Yihong},
  booktitle={Proceedings of the IEEE/CVF Conference on Computer Vision and Pattern Recognition},
  pages={9697--9706},
  year={2023}
}

@inproceedings{GRM,
  title={Generalized relation modeling for transformer tracking},
  author={Gao, Shenyuan and Zhou, Chunluan and Zhang, Jun},
  booktitle={Proceedings of the IEEE/CVF Conference on Computer Vision and Pattern Recognition},
  pages={18686--18695},
  year={2023}
}

@article{cui2024mixformerv2,
  title={Mixformerv2: Efficient fully transformer tracking},
  author={Cui, Yutao and Song, Tianhui and Wu, Gangshan and Wang, Limin},
  journal={Advances in Neural Information Processing Systems},
  volume={36},
  year={2024}
}

@article{kou2024zoomtrack,
  title={ZoomTrack: Target-aware Non-uniform Resizing for Efficient Visual Tracking},
  author={Kou, Yutong and Gao, Jin and Li, Bing and Wang, Gang and Hu, Weiming and Wang, Yizheng and Li, Liang},
  journal={Advances in Neural Information Processing Systems},
  volume={36},
  year={2024}
}

@inproceedings{odtrack,
  title={ODTrack: Online Dense Temporal Token Learning for Visual Tracking}, 
  author={Yaozong Zheng and Bineng Zhong and Qihua Liang and Zhiyi Mo and Shengping Zhang and Xianxian Li},
  booktitle={AAAI},
  year={2024}
}

@inproceedings{TaMOs,
  title={Beyond SOT: Tracking Multiple Generic Objects at Once},
  author={Mayer, Christoph and Danelljan, Martin and Yang, Ming-Hsuan and Ferrari, Vittorio and Van Gool, Luc and Kuznetsova, Alina},
  booktitle={Proceedings of the IEEE/CVF Winter Conference on Applications of Computer Vision},
  pages={6826--6836},
  year={2024}
}

@INPROCEEDINGS{OTB,
  author={Y. {Wu} and J. {Lim} and M. {Yang}},
  booktitle={2013 IEEE Conference on Computer Vision and Pattern Recognition}, 
  title={Online Object Tracking: A Benchmark}, 
  year={2013},
  volume={},
  number={},
  pages={2411-2418},
  doi={10.1109/CVPR.2013.312}}

@inproceedings{promptSOD,
  title={Explicit visual prompting for low-level structure segmentations},
  author={Liu, Weihuang and Shen, Xi and Pun, Chi-Man and Cun, Xiaodong},
  booktitle={Proceedings of the IEEE/CVF Conference on Computer Vision and Pattern Recognition},
  pages={19434--19445},
  year={2023}
}

@article{han2021novel,
  title={A novel occlusion-aware vote cost for light field depth estimation},
  author={Han, Kang and Xiang, Wei and Wang, Eric and Huang, Tao},
  journal={IEEE Transactions on Pattern Analysis and Machine Intelligence},
  volume={44},
  number={11},
  pages={8022--8035},
  year={2021},
  publisher={IEEE}
}

@article{li2023opal,
  title={Opal: Occlusion pattern aware loss for unsupervised light field disparity estimation},
  author={Li, Peng and Zhao, Jiayin and Wu, Jingyao and Deng, Chao and Han, Yuqi and Wang, Haoqian and Yu, Tao},
  journal={IEEE Transactions on Pattern Analysis and Machine Intelligence},
  year={2023},
  publisher={IEEE}
}

@article{wang2022disentangling,
  title={Disentangling light fields for super-resolution and disparity estimation},
  author={Wang, Yingqian and Wang, Longguang and Wu, Gaochang and Yang, Jungang and An, Wei and Yu, Jingyi and Guo, Yulan},
  journal={IEEE Transactions on Pattern Analysis and Machine Intelligence},
  volume={45},
  number={1},
  pages={425--443},
  year={2022},
  publisher={IEEE}
}

@inproceedings{strecke2017accurate,
  title={Accurate depth and normal maps from occlusion-aware focal stack symmetry},
  author={Strecke, Michael and Alperovich, Anna and Goldluecke, Bastian},
  booktitle={Proceedings of the IEEE Conference on Computer Vision and Pattern Recognition},
  pages={2814--2822},
  year={2017}
}

@article{srinivasan2017shape,
  title={Shape Estimation from Shading, Defocus, and Correspondence Using Light-Field Angular Coherence},
  author={Srinivasan, Michael W Tao Pratul P and Rusinkiewicz, Sunil Hadap Szymon and Ramamoorthi, Jitendra Malik Ravi},
  journal={IEEE Transactions on Pattern Analysis and Machine Intelligence},
  volume={39},
  number={3},
  year={2017}
}

@article{zhang2016light,
  title={Light-field depth estimation via epipolar plane image analysis and locally linear embedding},
  author={Zhang, Yongbing and Lv, Huijin and Liu, Yebin and Wang, Haoqian and Wang, Xingzheng and Huang, Qian and Xiang, Xinguang and Dai, Qionghai},
  journal={IEEE Transactions on Circuits and Systems for Video Technology},
  volume={27},
  number={4},
  pages={739--747},
  year={2016},
  publisher={IEEE}
}

@article{zhou2023beyond,
  title={Beyond Photometric Consistency: Geometry-Based Occlusion-Aware Unsupervised Light Field Disparity Estimation},
  author={Zhou, Wenhui and Lin, Lili and Hong, Yongjie and Li, Qiujian and Shen, Xingfa and Kuruoglu, Ercan Engin},
  journal={IEEE Transactions on Neural Networks and Learning Systems},
  year={2023},
  publisher={IEEE}
}

@inproceedings{wang2022lfbcnet,
  title={LFBCNet: Light Field Boundary-aware and Cascaded Interaction Network for Salient Object Detection},
  author={Wang, Mianzhao and Shi, Fan and Cheng, Xu and Zhao, Meng and Zhang, Yao and Jia, Chen and Tian, Weiwei and Chen, Shengyong},
  booktitle={Proceedings of the 30th ACM International Conference on Multimedia},
  pages={3430--3439},
  year={2022}
}

@article{zhang2021geometry,
  title={Geometry auxiliary salient object detection for light fields via graph neural networks},
  author={Zhang, Qiudan and Wang, Shiqi and Wang, Xu and Sun, Zhenhao and Kwong, Sam and Jiang, Jianmin},
  journal={IEEE Transactions on Image Processing},
  volume={30},
  pages={7578--7592},
  year={2021},
  publisher={IEEE}
}

@inproceedings{cong2023combining,
  title={Combining Implicit-Explicit View Correlation for Light Field Semantic Segmentation},
  author={Cong, Ruixuan and Yang, Da and Chen, Rongshan and Wang, Sizhe and Cui, Zhenglong and Sheng, Hao},
  booktitle={Proceedings of the IEEE/CVF Conference on Computer Vision and Pattern Recognition},
  pages={9172--9181},
  year={2023}
}

@article{wang2022visual,
  title={Visual Object Tracking Based on Light-Field Imaging in the Presence of Similar Distractors},
  author={Wang, Mianzhao and Shi, Fan and Cheng, Xu and Zhao, Meng and Zhang, Yao and Jia, Chen and Tian, Weiwei and Chen, Shengyong},
  journal={IEEE Transactions on Industrial Informatics},
  volume={19},
  number={3},
  pages={2705--2716},
  year={2022},
  publisher={IEEE}
}

@inproceedings{ECO,
  title={Eco: Efficient convolution operators for tracking},
  author={Danelljan, Martin and Bhat, Goutam and Shahbaz Khan, Fahad and Felsberg, Michael},
  booktitle={Proceedings of the IEEE conference on computer vision and pattern recognition},
  pages={6638--6646},
  year={2017}
}

@inproceedings{ATOM,
  title={Atom: Accurate tracking by overlap maximization},
  author={Danelljan, Martin and Bhat, Goutam and Khan, Fahad Shahbaz and Felsberg, Michael},
  booktitle={Proceedings of the IEEE/CVF conference on computer vision and pattern recognition},
  pages={4660--4669},
  year={2019}
}

@inproceedings{PrDiMP,
  title={Probabilistic regression for visual tracking},
  author={Danelljan, Martin and Gool, Luc Van and Timofte, Radu},
  booktitle={Proceedings of the IEEE/CVF conference on computer vision and pattern recognition},
  pages={7183--7192},
  year={2020}
}

@inproceedings{DiMP,
  title={Learning discriminative model prediction for tracking},
  author={Bhat, Goutam and Danelljan, Martin and Gool, Luc Van and Timofte, Radu},
  booktitle={Proceedings of the IEEE/CVF international conference on computer vision},
  pages={6182--6191},
  year={2019}
}

@inproceedings{li2019siamrpn++,
  title={Siamrpn++: Evolution of siamese visual tracking with very deep networks},
  author={Li, Bo and Wu, Wei and Wang, Qiang and Zhang, Fangyi and Xing, Junliang and Yan, Junjie},
  booktitle={Proceedings of the IEEE/CVF conference on computer vision and pattern recognition},
  pages={4282--4291},
  year={2019}
}

@inproceedings{zhang2019learning,
  title={Learning the model update for siamese trackers},
  author={Zhang, Lichao and Gonzalez-Garcia, Abel and Weijer, Joost Van De and Danelljan, Martin and Khan, Fahad Shahbaz},
  booktitle={Proceedings of the IEEE/CVF international conference on computer vision},
  pages={4010--4019},
  year={2019}
}

@inproceedings{SiamDW,
  title={Deeper and wider siamese networks for real-time visual tracking},
  author={Zhang, Zhipeng and Peng, Houwen},
  booktitle={Proceedings of the IEEE/CVF conference on computer vision and pattern recognition},
  pages={4591--4600},
  year={2019}
}

@article{wang2021unsupervised,
  title={Unsupervised deep representation learning for real-time tracking},
  author={Wang, Ning and Zhou, Wengang and Song, Yibing and Ma, Chao and Liu, Wei and Li, Houqiang},
  journal={International Journal of Computer Vision},
  volume={129},
  pages={400--418},
  year={2021},
  publisher={Springer}
}

@inproceedings{SiamFc,
  title={Fully-convolutional siamese networks for object tracking},
  author={Bertinetto, Luca and Valmadre, Jack and Henriques, Joao F and Vedaldi, Andrea and Torr, Philip HS},
  booktitle={Computer Vision--ECCV 2016 Workshops: Amsterdam, The Netherlands, October 8-10 and 15-16, 2016, Proceedings, Part II 14},
  pages={850--865},
  year={2016},
  organization={Springer}
}

@inproceedings{OSTrack,
  title={Joint feature learning and relation modeling for tracking: A one-stream framework},
  author={Ye, Botao and Chang, Hong and Ma, Bingpeng and Shan, Shiguang and Chen, Xilin},
  booktitle={European Conference on Computer Vision},
  pages={341--357},
  year={2022},
  organization={Springer}
}

@inproceedings{MixFormer,
  title={Mixformer: End-to-end tracking with iterative mixed attention},
  author={Cui, Yutao and Jiang, Cheng and Wang, Limin and Wu, Gangshan},
  booktitle={Proceedings of the IEEE/CVF Conference on Computer Vision and Pattern Recognition},
  pages={13608--13618},
  year={2022}
}

@inproceedings{STARK,
  title={Learning spatio-temporal transformer for visual tracking},
  author={Yan, Bin and Peng, Houwen and Fu, Jianlong and Wang, Dong and Lu, Huchuan},
  booktitle={Proceedings of the IEEE/CVF international conference on computer vision},
  pages={10448--10457},
  year={2021}
}

@article{SwinTrack,
  title={Swintrack: A simple and strong baseline for transformer tracking},
  author={Lin, Liting and Fan, Heng and Zhang, Zhipeng and Xu, Yong and Ling, Haibin},
  journal={Advances in Neural Information Processing Systems},
  volume={35},
  pages={16743--16754},
  year={2022}
}

@ARTICLE{Survey_self_Supervised_1,
  author={Wang, Jiangliu and Jiao, Jianbo and Bao, Linchao and He, Shengfeng and Liu, Wei and Liu, Yun-hui},
  journal={IEEE Transactions on Pattern Analysis and Machine Intelligence}, 
  title={Self-Supervised Video Representation Learning by Uncovering Spatio-Temporal Statistics}, 
  year={2022},
  volume={44},
  number={7},
  pages={3791-3806},
  doi={10.1109/TPAMI.2021.3057833}}

@ARTICLE{Survey_self_Supervised_5,
  author={Jing, Longlong and Tian, Yingli},
  journal={IEEE Transactions on Pattern Analysis and Machine Intelligence}, 
  title={Self-Supervised Visual Feature Learning With Deep Neural Networks: A Survey}, 
  year={2021},
  volume={43},
  number={11},
  pages={4037-4058},
  doi={10.1109/TPAMI.2020.2992393}}

@inproceedings{Devlin2019BERTPO,
  title={BERT: Pre-training of Deep Bidirectional Transformers for Language Understanding},
  author={Jacob Devlin and Ming-Wei Chang and Kenton Lee and Kristina Toutanova},
  booktitle={North American Chapter of the Association for Computational Linguistics},
  year={2019},
  url={https://api.semanticscholar.org/CorpusID:52967399}
}

@INPROCEEDINGS{MAE,
  author={He, Kaiming and Chen, Xinlei and Xie, Saining and Li, Yanghao and Dollár, Piotr and Girshick, Ross},
  booktitle={2022 IEEE/CVF Conference on Computer Vision and Pattern Recognition (CVPR)}, 
  title={Masked Autoencoders Are Scalable Vision Learners}, 
  year={2022},
  volume={},
  number={},
  pages={15979-15988},
  doi={10.1109/CVPR52688.2022.01553}}

@inproceedings{vondrick2018tracking,
  title={Tracking emerges by colorizing videos},
  author={Vondrick, Carl and Shrivastava, Abhinav and Fathi, Alireza and Guadarrama, Sergio and Murphy, Kevin},
  booktitle={Proceedings of the European conference on computer vision (ECCV)},
  pages={391--408},
  year={2018}
}

@inproceedings{wu2023dropmae,
  title={DropMAE: Masked Autoencoders with Spatial-Attention Dropout for Tracking Tasks},
  author={Wu, Qiangqiang and Yang, Tianyu and Liu, Ziquan and Wu, Baoyuan and Shan, Ying and Chan, Antoni B},
  booktitle={Proceedings of the IEEE/CVF Conference on Computer Vision and Pattern Recognition},
  pages={14561--14571},
  year={2023}
}

@inproceedings{sun2023masked,
  title={Masked Motion Encoding for Self-Supervised Video Representation Learning},
  author={Sun, Xinyu and Chen, Peihao and Chen, Liangwei and Li, Changhao and Li, Thomas H and Tan, Mingkui and Gan, Chuang},
  booktitle={Proceedings of the IEEE/CVF Conference on Computer Vision and Pattern Recognition},
  pages={2235--2245},
  year={2023}
}

@article{gupta2023siamese,
  title={Siamese Masked Autoencoders},
  author={Gupta, Agrim and Wu, Jiajun and Deng, Jia and Fei-Fei, Li},
  journal={arXiv preprint arXiv:2305.14344},
  year={2023}
}

@inproceedings{MCITrack,
  title={Exploring enhanced contextual information for video-level object tracking},
  author={Kang, Ben and Chen, Xin and Lai, Simiao and Liu, Yang and Liu, Yi and Wang, Dong},
  booktitle={Proceedings of the AAAI Conference on Artificial Intelligence},
  volume={39},
  number={4},
  pages={4194--4202},
  year={2025}
}

@article{SparseTrack,
  title={Sparsetrack: Multi-object tracking by performing scene decomposition based on pseudo-depth},
  author={Liu, Zelin and Wang, Xinggang and Wang, Cheng and Liu, Wenyu and Bai, Xiang},
  journal={IEEE Transactions on Circuits and Systems for Video Technology},
  year={2025},
  publisher={IEEE}
}

@inproceedings{Sutrack,
  title={Sutrack: Towards simple and unified single object tracking},
  author={Chen, Xin and Kang, Ben and Geng, Wanting and Zhu, Jiawen and Liu, Yi and Wang, Dong and Lu, Huchuan},
  booktitle={Proceedings of the AAAI Conference on Artificial Intelligence},
  volume={39},
  number={2},
  pages={2239--2247},
  year={2025}
}

\end{document}